\documentclass[10pt, a4paper]{article}

\usepackage[final]{lrec2026} 

\usepackage{graphicx}
\usepackage[table]{xcolor}
\usepackage{inconsolata}
\usepackage{microtype}
\usepackage{booktabs}

\usepackage{tabularx}
\usepackage[utf8]{inputenc}
\usepackage{multirow}

\usepackage{amssymb}
\usepackage{pifont}
\usepackage{tikz} 
\usepackage[T1]{fontenc}
\usepackage{amsmath}

\title{Multimodal Task Interference: A Benchmark and Analysis of History-Target Mismatch in Multimodal LLMs}

\name{Masayuki Kawarada, Tatsuya Ishigaki, Hiroya Takamura} 

\address{Artificial Intelligence Research Center, AIST \\
         \texttt{\{kawarada.masayuki, ishigaki.tatsuya, takamura.hiroya\}@aist.go.jp}\\}

\abstract{
\emph{Task interference}, the performance degradation caused by task switches within a single conversation, has been studied exclusively in text-only settings despite the growing prevalence of multimodal dialogue systems. We introduce a benchmark for evaluating this phenomenon in multimodal LLMs, covering six tasks across text and vision with systematic variation of history-target along three axes: modality mismatch, reasoning mismatch, and answer format mismatch.
Experiments on both open-weights and proprietary models reveal that task interference is highly directional: switching from text-only to image-based targets causes severe performance drops, while the reverse transition yields minimal degradation. Interference is further amplified when mismatches co-occur across multiple dimensions, and is driven most strongly by modality differences, followed by answer format, while reasoning requirement shifts cause minimal degradation.
\\ \newline 
}

\begin{document}

\maketitleabstract

\section{Introduction}
Large language models (LLMs) have achieved strong performance across a wide range of tasks, including question answering, image captioning, and sentiment classification~\citep{brown2020language, openai2023gpt4}.
In real-world dialogue scenarios, users frequently perform multiple tasks in succession within a single conversation.
When the input transitions from one task (e.g., image captioning) to another (e.g., textual question answering), a phenomenon known as \emph{task switching} occurs (see Figure 1 for an illustrative example). Such transitions can lead to a decline in model performance, an effect termed \emph{task interference}~\citep{gupta2024llm}.

Although task interference has been documented in text-only settings, existing studies do not account for the multimodal nature of modern dialogue systems, which increasingly involve both text and images.
This gap motivates the need for a dedicated evaluation framework that can systematically quantify how different types of mismatches between dialogue history and target inputs affect model performance. Beyond modality considerations alone, we hypothesize that the compatibility between reasoning requirements and answer format~(e.g., classification versus generation) also plays a significant role in determining the degree of interference.

To address this need, we introduce a benchmark designed to evaluate task interference in multimodal LLMs. Our benchmark is built around three central research questions.
First, how does a modality mismatch between history and target~(e.g., transitioning from image-based to text-based input) affect model performance? 
Second, how does a reasoning mismatch, such as switching from commonsense question answering to factual recall, influence accuracy? 
Third, how does an answer format mismatch, such as transitioning from classification to generation, affect output quality?

\begin{figure}[t]
    \centering
    \includegraphics[width=0.95\linewidth]{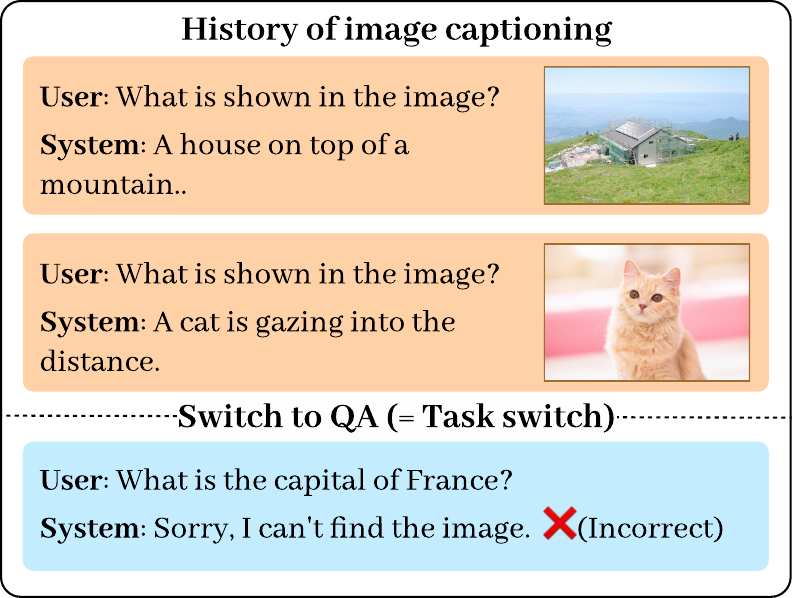}
    \caption{An illustrative example of multimodal task interference. After processing a conversational history of image captioning tasks, the model is prompted with a text-only question. The sudden task switch causes the model to erroneously expect a visual input, leading to a failure in answering a simple factual question.}
    \label{fig:task_switch}
\end{figure}

\begin{table*}[t]
    \centering
    \small
    \begin{tabular}{lcccc}
        \toprule
        \textbf{Dataset} & \textbf{Task Type} & \textbf{Modality ($\mathcal{M}$)} & \textbf{Reasoning ($\mathcal{R}$)} & \textbf{Answer Format ($\mathcal{A}$)} \\
        \midrule
        Rotten Tomatoes & Sentiment Classification & Text-only & No & Classification \\
        MMLU & Multiple-choice QA & Text-only & Yes & Multiple-choice \\
        TweetQA & Open-ended QA & Text-only & No & Generation \\
        \midrule
        VQAv2 & Visual QA & Image + Text & No & Short-answer \\
        OK-VQA & Visual QA & Image + Text & Yes & Short-answer \\
        COCO Captions & Image Captioning & Image + Text & No & Generation \\
        \bottomrule
    \end{tabular}
    \caption{Overview of the six datasets used in our benchmark, categorized by the three axes of our task interference framework: Modality~($\mathcal{M}$), Reasoning requirement~($\mathcal{R}$), and Answer Format~($\mathcal{A}$).}
    \label{tab:task_summary}
\end{table*}

The benchmark encompasses six diverse datasets covering sentiment classification, multiple-choice question answering, open-ended question answering, image captioning, and visual question answering. By constructing varied history-target configurations across these datasets, we enable controlled analysis of each mismatch dimension. We apply this benchmark to both open-weights and proprietary multimodal LLMs, allowing a direct comparison between the two categories under a unified evaluation protocol.

Our results reveal that task interference in multimodal LLMs is highly directional and multifaceted. Cross-modal transitions exhibit a stark asymmetry: switching from text-only histories to image-based targets causes catastrophic performance drops, while the reverse yields minimal interference. Furthermore, simultaneous mismatches across multiple dimensions compound this degradation, demonstrating that modality mismatch alone cannot fully explain the interference. Models also show susceptibility to answer format changes but unexpected robustness to shifts in reasoning, highlighting the crucial role of structural and cognitive compatibility in dialogue stability.

Our contributions are: (1) a benchmark for evaluating task interference in multimodal LLMs along three axes (modality, reasoning, and answer format mismatch); (2) a comprehensive empirical study across six datasets with both open-weights and proprietary models; and (3) evidence that while modality switches cause significant interference, the performance degradation is most severe when compounded by simultaneous mismatches in reasoning requirements and answer formats.

\section{Related Work}

\subsection{Multimodal Large Language Models}
Multimodal large language models (MLLMs) extend LLMs to handle inputs across text and image modalities.
Pioneering work such as Flamingo~\cite{alayrac2022flamingo} demonstrated few-shot learning over interleaved image--text sequences, and subsequent systems including LLaVA~\cite{visual_instruction_tuning}, GPT-4~\cite{openai2023gpt4}, and Qwen3-VL~\cite{bai2025qwen3vltechnicalreport} have further advanced vision-language alignment and general-purpose assistant capabilities.
As these models are increasingly deployed in multi-turn dialogue settings, the effect of accumulating heterogeneous conversational history becomes a practical concern.
Prior work on long-context LLMs has shown that performance degrades non-uniformly as context grows, with information in the middle of long inputs being systematically overlooked~\cite{liu-etal-2024-lost}.
In multimodal dialogues, this challenge is further compounded by cross-modal history, motivating dedicated analysis of how different history compositions affect model behavior.

\subsection{Task Interference}
Task interference in LLMs has been studied both as a problem to mitigate and as a phenomenon to analyze.
From the mitigation side, \citet{chen2023octavius} and \citet{shen2024multimodal} propose mixture-of-LoRA architectures to reduce cross-task conflicts in MLLMs.
From the analysis side, \citet{gupta2024llm} formally define task interference as the performance degradation caused by task-switched conversational history in text-only settings, demonstrating significant accuracy drops across multiple task configurations.
However, their analysis is limited to unimodal text settings and does not account for modality switches.
Our work extends this line of research to multimodal dialogue by systematically evaluating interference along three axes of modality, reasoning requirement, and answer format, revealing interference patterns that text-only studies cannot capture, particularly a stark asymmetry in cross-modal transitions.

\section{Task Interference}
\label{sec:task_interference}

We formalize task interference in MLLMs. Let $f$ be a model evaluated on a target task $T_{\text{tgt}}$ with input $x_{\text{tgt}}$ and reference $y_{\text{tgt}}$. In a conversational setting, the model is conditioned on a dialogue history $H = \{(x_i, y_i)\}_{i=1}^N$ of length $N$.

To systematically isolate the effect of \emph{task switching}, we distinguish between two types of dialogue history:
\begin{itemize}
    \item \textbf{Same-task History ($H_{\text{same}}$)}: The history consists of examples from the target task itself ($T_H = T_{\text{tgt}}$), effectively acting as in-context learning.
    \item \textbf{Switched-task History ($H_{\text{switch}}$)}: The history is sampled from a different task ($T_H \neq T_{\text{tgt}}$), introducing a task switch.
\end{itemize}

We quantify \emph{Task Interference}, $\Delta_{\text{switch}}^{(\%)}$, as the relative performance degradation caused by a task switch compared to the ideal scenario where the context is consistent with the target task. Given an evaluation metric $E$, let $E_{\text{switch}}$ and $E_{\text{same}}$ be the performance scores under the switched-task and same-task histories, respectively:
\begin{align*}
    E_{\text{switch}} &= E(f(H_{\text{switch}}, x_{\text{tgt}}), y_{\text{tgt}}), \\
    E_{\text{same}} &= E(f(H_{\text{same}}, x_{\text{tgt}}), y_{\text{tgt}}),
\end{align*}
where $f(H, x_{\text{tgt}})$ denotes the model output when conditioned on the dialogue history $H$ followed by the target input $x_{\text{tgt}}$. We then define the switch cost as:
$$ \Delta_{\text{switch}}^{(\%)} = 100 \cdot \frac{E_{\text{switch}} - E_{\text{same}}}{E_{\text{same}}}$$

Unlike prior studies \citep{gupta2024llm} that measure interference against a zero-shot baseline ($H=\emptyset$), our formulation isolates the specific impact of the \emph{task switch}. By holding the presence of a conversation history constant, we control for confounding factors inherent to conversational prompting, such as increased context length, general in-context learning dynamics, and susceptibility to format failures. Task interference occurs when $\Delta_{\text{switch}}^{(\%)} < 0$, indicating that a switched-task history degrades performance compared to a relevant, same-task history.

To analyze the drivers of this interference, we characterize any task $T$ as a tuple $T = \langle \mathcal{M}, \mathcal{R}, \mathcal{A} \rangle$, representing modality (e.g., text-only, image+text), reasoning requirement (e.g., factual recall, commonsense), and answer format (e.g., classification, generation). While a task switch occurs whenever the overall task changes ($T_H \neq T_{\text{tgt}}$), the specific attributes between the history and target tasks can independently match or differ. 
Therefore, within switched-task scenarios, we define the relationship along each axis as either a match~(the attribute remains identical) or a mismatch~(the attribute differs). 
Our benchmark evaluates interference by comparing these matched and mismatched conditions across three specific dimensions: modality ($\mathcal{M}_H = \mathcal{M}_{\text{tgt}}$ vs. $\mathcal{M}_H \neq \mathcal{M}_{\text{tgt}}$), reasoning ($\mathcal{R}_H = \mathcal{R}_{\text{tgt}}$ vs. $\mathcal{R}_H \neq \mathcal{R}_{\text{tgt}}$), and answer format ($\mathcal{A}_H = \mathcal{A}_{\text{tgt}}$ vs. $\mathcal{A}_H \neq \mathcal{A}_{\text{tgt}}$).

\section{Experiments}

We systematically evaluate model performance across all history-target, as detailed in the next section.

\subsection{Target Tasks and Datasets}

We use six benchmark datasets as shown in Table~\ref{tab:task_summary}, spanning diverse task types and input modalities. 
Our dataset selection is guided by three criteria:
(1) covering both text and image modalities to evaluate modality-specific and cross-modal interference,
(2) controlling task difficulty by including tasks that require commonsense reasoning versus those that rely mainly on surface-level understanding, and
(3) incorporating both classification/QA tasks with clear-cut answers and generation tasks where the output is inherently open-ended.

Table~\ref{tab:task_summary} shows the datasets classified in terms of these criteria.
For text-based tasks, we use \textbf{Rotten Tomatoes}~\cite{pang2005} for binary sentiment classification (commonsense not required), \textbf{Massive Multitask Language Understanding (MMLU)}~\cite{hendrycks2021} for multiple-choice question answering in the mathematics domain (requiring multi-step reasoning), and \textbf{TweetQA}~\cite{xiong2019tweetqa} for open-ended QA over social media posts (commonsense not required).
For image-based tasks, we include \textbf{OK-VQA}~\cite{marino2019okvqa}, which requires external commonsense knowledge beyond visual content, \textbf{VQAv2}~\cite{goyal2017vqa}, which focuses on object recognition with minimal reasoning, and \textbf{COCO Captions}~\cite{lin2014coco} for open-ended image caption generation.
We examine LLM performance on all combinations of these datasets.

\subsection{Multimodal Large Language Models}

To comprehensively assess task interference, we evaluate four representative multimodal large language models, encompassing both proprietary API-based and open-weights architectures. 

For the proprietary model, we evaluate \textbf{GPT-4.1-mini}\footnote{\texttt{gpt-4.1-mini-2025-04-14}}~\cite{openai2023gpt4}. For the open-weights models, we evaluate \textbf{Gemma-3n}\footnote{\texttt{google/gemma-3n-E4B-it}}~\cite{gemmateam2025gemma3technicalreport}, \textbf{Qwen3-VL}\footnote{\texttt{Qwen/Qwen3-VL-30B-A3B-Instruct}}~\cite{yang2025qwen3technicalreport}, and \textbf{Pixtral}\footnote{\texttt{mistralai/Pixtral-12B-2409}}~\cite{agrawal2024pixtral12b}. By including models with varying parameter scales and fusion mechanisms, we aim to determine whether susceptibility to task interference is a universal characteristic of multimodal systems or highly model-dependent.

\begin{table*}[t]
\centering
\small
\resizebox{\textwidth}{!}{
\begin{tabular}{lccc ccc ccc}
\toprule

& \multicolumn{3}{c}{\textbf{Modality (\%)}}
& \multicolumn{3}{c}{\textbf{Reasoning (\%)}}
& \multicolumn{3}{c}{\textbf{Answer Format (\%)}} \\
\cmidrule(lr){2-4}\cmidrule(lr){5-7}\cmidrule(lr){8-10}
& mismatch & match & $\Delta$
& mismatch & match & $\Delta$
& mismatch & match & $\Delta$ \\
\midrule
\multicolumn{10}{l}{\textbf{N=1}} \\
\hspace{2mm}GPT-4.1-mini & -8.89 & -7.65 & -1.24 & -5.35 & -11.88 & 6.53 & -8.44 & -8.12 & -0.32 \\
\hspace{2mm}Gemma-3n     & -10.49 & -11.48 & 0.99 & -10.84 & -10.94 & 0.10 & -11.11 & -9.39 & -1.72 \\
\hspace{2mm}Qwen3-VL     & -9.50 & 2.65 & -12.15$^{***}$ & -4.29 & -5.03 & 0.74 & -5.35 & -0.03 & -5.31$^{**}$ \\
\hspace{2mm}Pixtral      & -6.15 & -1.34 & -4.81$^{***}$ & -4.05 & -4.42 & 0.38 & -3.83 & -6.75 & 2.91 \\
\midrule
\multicolumn{10}{l}{\textbf{N=3}} \\
\hspace{2mm}GPT-4.1-mini & -16.63 & -11.55 & -5.09$^{**}$ & -13.49 & -15.87 & 2.38 & -15.41 & -9.33 & -6.08$^{**}$ \\
\hspace{2mm}Gemma-3n     & -22.79 & -20.50 & -2.29$^{*}$ & -22.05 & -21.67 & -0.38 & -21.94 & -21.43 & -0.52 \\
\hspace{2mm}Qwen3-VL     & -14.58 & -1.02 & -13.56$^{***}$ & -7.74 & -10.78 & 3.05 & -8.96 & -10.41 & 1.45 \\
\hspace{2mm}Pixtral      & -8.54 & -2.22 & -6.31$^{***}$ & -5.17 & -6.97 & 1.80 & -5.66 & -8.30 & 2.64 \\
\midrule
\multicolumn{10}{l}{\textbf{N=5}} \\
\hspace{2mm}GPT-4.1-mini & -19.99 & -14.02 & -5.97$^{**}$ & -16.25 & -19.15 & 2.91 & -18.48 & -11.90 & -6.58$^{**}$ \\
\hspace{2mm}Gemma-3n     & -25.25 & -21.66 & -3.60$^{*}$ & -23.68 & -23.97 & 0.29 & -24.25 & -21.02 & -3.23 \\
\hspace{2mm}Qwen3-VL     & -14.87 & -3.18 & -11.69$^{***}$ & -8.93 & -11.63 & 2.70 & -10.45 & -8.52 & -1.93 \\
\hspace{2mm}Pixtral      & -11.09 & -4.39 & -6.69$^{***}$ & -8.14 & -8.72 & 0.58 & -8.09 & -10.46 & 2.37 \\
\bottomrule
\end{tabular}
}
\caption{Axis-wise means (\%) for mismatch and match groups, and their difference ($\Delta$ = mismatch - match). Significance marks on $\Delta$: Welch's t-test ($^{*}p<0.05,\,^{**}p<0.01,\,^{***}p<0.001$).}
\label{tab:model_axis_same_switch_delta}
\end{table*}

\subsection{Experimental Setup}
For all evaluated models, the input prompts contain $N=1, 3, 5$ randomly sampled history examples followed by a target input. For tasks involving images, visual inputs are passed through each model's native multimodal interface by substituting image-slot tokens in the prompt template with base64-encoded image representations.

To investigate the precise impact of a task switch, we adopt a teacher-forcing approach when constructing the history prompts. Specifically, the response for each history example is taken directly from the ground-truth label or reference text. 
This methodology ensures that the dialogue context is perfectly accurate and eliminates the confounding factor of model generation errors propagating through the conversation. 

To ensure the reproducibility and statistical robustness of our results, 
all experiments are conducted across five different random seeds for each condition. For each seed, we resample the history examples to account for variance in in-context example selection. Throughout these trials, the decoding temperature is set to 0 to eliminate stochasticity in model outputs, ensuring that observed variance reflects only the effect of history composition.

For the open-weights models, all local inference was conducted on a single NVIDIA H200 GPU using the vLLM inference framework~\cite{kwon2023efficient}.
For evaluation, we use the initial 1,000 instances from the test split of each target dataset.

\subsection{Evaluation Metrics}
To assess model performance, we select primary evaluation metrics tailored to the specific output format of each dataset. 
For classification and multiple-choice question answering tasks, specifically Rotten Tomatoes and MMLU, we report standard accuracy. 
For open-ended textual question answering on TweetQA, we evaluate the responses using the F1 score to measure the token-level overlap with the reference answers. 

For visual question answering tasks, encompassing VQAv2 and OK-VQA, we employ the standard VQA accuracy metric to account for human annotator variance. 
Finally, for the open-ended image captioning task on COCO Captions, we use the CIDEr metric~\cite{Vedantam_2015_CVPR}, which effectively evaluates the consensus between the generated caption and human reference captions. 

Since the absolute scales of these metrics vary significantly (e.g., CIDEr scores versus standard percentages), we evaluate the switch cost using the relative percentage change ($\Delta_{\text{switch}}^{(\%)}$) as defined in Section~\ref{sec:task_interference}. This normalization enables a unified and fair comparison of task interference across all datasets and diverse metric types.

\section{Results and Discussion}
\label{sec:results}

We organize our findings according to three hypothesized factors contributing to task interference: (1) modality mismatch, (2) reasoning mismatch, and (3) answer format mismatch.

\subsection{Effects of Task Interference along Three Axes}
\label{sec:main_results}
Table~\ref{tab:model_axis_same_switch_delta} presents the performance changes across the three mismatch axes.
We observe a consistent trend where performance generally degrades even in the ``match'' conditions. This demonstrates that the mere occurrence of a task switch, shifting from one dataset to another while preserving the same modality, reasoning requirement, or answer format, inherently harms model accuracy. This baseline degradation strongly aligns with prior findings on task interference in text-only dialogue settings \citep{gupta2024llm}. Building upon this observation, we dissect the additional interference induced specifically by mismatches along each of our three proposed axes.

\paragraph{Modality Mismatch}
Modality mismatch generally induces severe task interference, often being the most dominant factor across the evaluated models. While the impact varies in the single-shot setting ($N=1$), increasing the history length to $N=3$ and $N=5$ reveals significant performance degradation for all models. For instance, Qwen3-VL experiences a massive drop with a $\Delta$ of -12.15\% even at $N=1$. Similarly, as the context lengthens to $N=5$, GPT-4.1-mini and Gemma-3n show significant decreases of -5.97\% and -3.60\%, respectively. These results suggest that multimodal models struggle heavily to adapt when the input modality shifts between the dialogue history and the target prompt.

\begin{figure}[t]
    \centering
    \includegraphics[width=\linewidth]{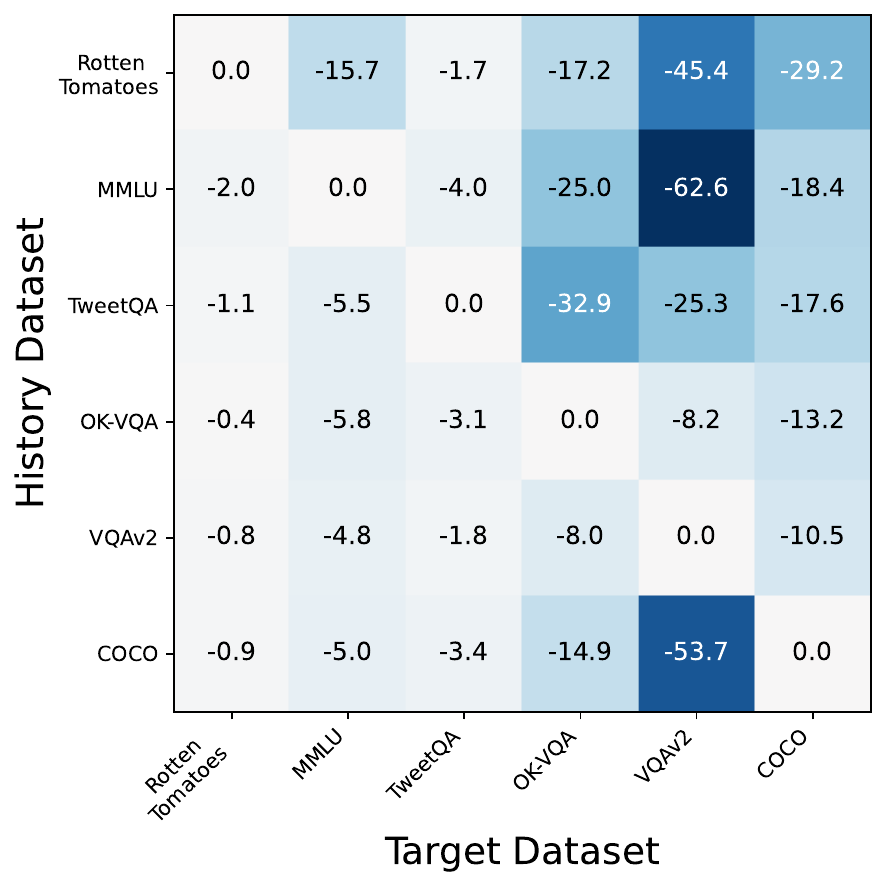}
    \caption{A heatmap visualizing the performance drop (relative change in \%) across all pairwise combinations of history and target datasets for GPT-4.1-mini with a history length of $N=3$.}
    \label{fig:delta_switch_heatmap}
\end{figure}

\paragraph{Reasoning Mismatch}
Interestingly, the data indicates that models are highly robust to shifts in reasoning requirements. Across almost all models and history lengths, the $\Delta$ values for reasoning mismatch are positive or near zero, and none reach statistical significance for negative degradation. For example, GPT-4.1-mini demonstrates a positive $\Delta$ of 6.53\% at $N=1$. This implies that switching between different cognitive tasks does not penalize model performance. In fact, maintaining the exact same reasoning type in the history might occasionally lead to over-conditioning on specific patterns, resulting in slightly worse outcomes than a mismatched history.

\paragraph{Answer Format Mismatch}
The effect of answer format mismatch is evident but highly model-dependent. At $N=1$, only Qwen3-VL shows a statistically significant drop of -5.31\%. However, as the dialogue context grows, GPT-4.1-mini becomes notably susceptible to format changes, exhibiting significant performance drops of -6.08\% at $N=3$ and -6.58\% at $N=5$. Conversely, open-weights models like Pixtral and Gemma-3n remain relatively unaffected by shifts in the expected answer format across all context lengths. This indicates that while format interference exists, it does not universally degrade performance in the same destructive manner as modality switching.

\begin{table}[t]
\centering
\small
\setlength{\tabcolsep}{3pt}
\begin{tabular}{lcc}
\toprule
 & \textbf{Text$\rightarrow$Image~(\%)} & \textbf{Image$\rightarrow$Text~(\%)} \\
\midrule
\multicolumn{3}{l}{\textbf{N=1}} \\
\hspace{2mm}GPT-4.1-mini & -18.65$^{***}$ & 0.88 \\
\hspace{2mm}Gemma-3n     & -18.87$^{***}$ & -2.11$^{***}$ \\
\hspace{2mm}Qwen3-VL     & -19.84$^{***}$ & 0.85 \\
\hspace{2mm}Pixtral      & -12.94$^{***}$ & 0.64 \\
\midrule
\multicolumn{3}{l}{\textbf{N=3}} \\
\hspace{2mm}GPT-4.1-mini & -30.38$^{***}$ & -2.88$^{***}$ \\
\hspace{2mm}Gemma-3n     & -39.91$^{***}$ & -5.67$^{***}$ \\
\hspace{2mm}Qwen3-VL     & -29.44$^{***}$ & 0.28 \\
\hspace{2mm}Pixtral      & -16.08$^{***}$ & -0.99 \\
\midrule
\multicolumn{3}{l}{\textbf{N=5}} \\
\hspace{2mm}GPT-4.1-mini & -35.48$^{***}$ & -4.50$^{***}$ \\
\hspace{2mm}Gemma-3n     & -42.70$^{***}$ & -7.81$^{***}$ \\
\hspace{2mm}Qwen3-VL     & -28.96$^{***}$ & -0.78$^{***}$ \\
\hspace{2mm}Pixtral      & -19.02$^{***}$ & -3.15$^{***}$ \\
\bottomrule
\end{tabular}
\caption{Directional modality transitions. Values are mean relative change (\%) for each transition type. Significance markers are based on one-sample t-test against 0 for each transition (\(^{*}p<0.05,\,^{**}p<0.01,\,^{***}p<0.001\)).}
\label{tab:modality_directional_transitions}
\end{table}

\begin{figure*}[t]
    \centering
    \includegraphics[width=\linewidth]{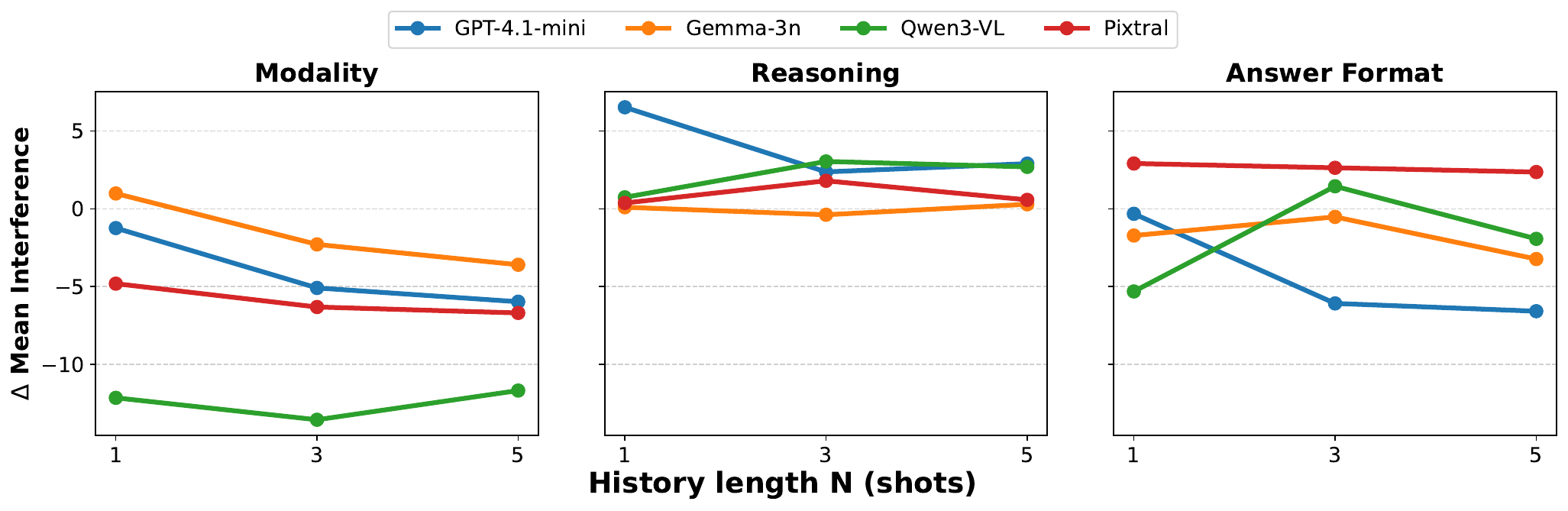}
    \caption{Performance difference ($\Delta$) between mismatch and match conditions across varying history lengths ($N=1, 3, 5$) for modality, reasoning, and answer format dimensions.}
    \label{fig:history_length}
\end{figure*}

\subsection{Interaction of Multiple Mismatches}
To better understand how the different dimensions of task interference interact, Figure~\ref{fig:delta_switch_heatmap} visualizes the performance drop across all pairwise history and target dataset combinations for GPT-4.1-mini at $N=3$.

First, cross-modal switches show a strong asymmetry. Transitions from text-only history to image-based targets suffer severe degradation, peaking at a 62.6\% drop from MMLU to VQAv2. Conversely, switching from image-based history to text-only targets yields marginal degradation, such as a mere 4.8\% drop from VQAv2 to MMLU. 
Second, simultaneous mismatches across multiple dimensions appear to contribute to a compound interference effect. The massive drop from MMLU to VQAv2 is likely exacerbated by the fact that the modality, reasoning, and answer format all change at once. 
Furthermore, even within the same modality, changing the reasoning and format requirements can be associated with significant interference, as seen in the 15.7\% drop from Rotten Tomatoes to MMLU.

\subsection{Asymmetry in Modality Transitions}
While Section~\ref{sec:main_results} established that modality mismatch significantly degrades overall performance, a closer examination reveals that this interference is highly directional. As shown in Table~\ref{tab:modality_directional_transitions}, there is a stark asymmetry between transitioning from text to image tasks and the reverse direction.

When models are conditioned on a text-only dialogue history and then prompted with a target image+text input, they suffer catastrophic performance drops. For example, as the context length increases to $N=5$, the Text$\rightarrow$Image transition results in massive decreases ranging from -19.02\% in Pixtral to -42.70\% in Gemma-3n. All evaluated models show highly statistically significant degradation in this specific direction.

Conversely, switching from an image-based history to a purely textual target task yields minimal interference. In the single-shot setting ($N=1$), models like GPT-4.1-mini and Qwen3-VL exhibit slightly positive relative changes of 0.88\% and 0.85\%, respectively. Even at the longest context length ($N=5$), the degradation in the Image$\rightarrow$Text direction remains mostly in the single digits. This strong asymmetry suggests that accumulating a long textual context overwhelms the visual processing capabilities of the models. They appear to lock into a text-only reasoning mode, causing them to neglect or severely misinterpret the newly introduced visual input.

\subsection{Impact of History Length}
To understand how the accumulation of context affects task switching, Figure~\ref{fig:history_length} illustrates the performance difference ($\Delta$) across varying history lengths ($N=1, 3, 5$) for each mismatch dimension. The trajectories reveal that the effect of history length is highly dependent on the type of mismatch.

In the case of modality mismatch, extending the history length generally exacerbates the interference. For models like GPT-4.1-mini, Gemma-3n, and Pixtral, the $\Delta$ becomes increasingly negative as the number of shots increases from $N=1$ to $N=5$. This steady downward trend indicates that a longer exposure to a specific modality strongly anchors the model's attention, making it progressively more difficult to process a sudden cross-modal target.

Conversely, the reasoning dimension exhibits remarkable stability across varying context lengths. The performance curves remain relatively flat and hover near or above the zero mark for all evaluated models. This visual evidence confirms that accumulating more examples of a mismatched reasoning type does not compound the cognitive interference, allowing the models to seamlessly adapt to the target prompt's reasoning requirement regardless of the context length.

Finally, the answer format dimension shows model-specific sensitivities to history length. While open-weights models maintain a relatively stable performance difference across varying $N$, GPT-4.1-mini displays a sharp performance drop as the context expands from $N=1$ to $N=3$. This suggests that certain models become easily locked into the answer format established by a longer dialogue history, severely hindering their ability to adapt to a different target format.

\subsection{Qualitative Analysis of Task Interference Mechanisms}
To qualitatively understand the interference mechanism, we manually analyzed the generated errors and observed a specific type of output-style drift induced by task-switched history.
We observe clear output-style drift when visual target tasks are preceded by text-only history. For short-answer visual QA (VQAv2/OK-VQA), while same-task visual history leads to concise, task-appropriate answers, switched-task text-only history often yields verbose or reformulated outputs that deviate from the expected gold labels.

For example, in OK-VQA, the same-task output is a concise ``soccer'', but MMLU history switches the model's response to the more descriptive but non-matching ``playing catch.''
Similarly, the same-task answer ``ball'' shifts to ``frisbee'' under Rotten Tomatoes history.
Even when the semantics are largely preserved, formatting drift can lead to evaluation failures: in VQAv2, the same-task output is a simple ``yes'', while MMLU history induces the redundant sentence ``Yes, the leaves are large.'' 

Quantitatively, this bias is reflected in the average output length for VQAv2 targets, which increases from 1.69 words under same-task history to 3.69 words under MMLU history for GPT-4.1-mini.
These results suggest that long textual contexts anchor the model into an elaborative completion mode, which is fundamentally incompatible with constrained multimodal evaluation metrics.

\section{Conclusion}
This paper investigated task interference in multimodal large language models. Our evaluation reveals that performance degradation is highly directional, exhibiting a stark asymmetry. Transitioning from text-only histories to image-based targets severely degrades performance, whereas the reverse causes minimal disruption.
Furthermore, task interference compounds when models face simultaneous mismatches across multiple dimensions. The most catastrophic drops occur when modality, reasoning, and answer format change at once. Interestingly, models are vulnerable to modality and answer format shifts but remain unexpectedly robust to changes in reasoning requirements. 

These findings demonstrate that modality matching alone cannot guarantee conversational stability. To build truly robust multimodal dialogue systems, future work must focus on dynamic context management, interference detection, and mixed-task instruction tuning.

\section*{Limitations}
Our study has several limitations that highlight directions for future work. First, while we evaluate four representative MLLMs, our analysis does not cover the largest flagship models (e.g., full-scale GPT-4.1 or 70B+ parameter models). Investigating how model scale affects robustness remains an important next step.

Second, our experiments focus on short dialogue histories ($N \le 5$) with a single task switch, whereas real-world conversations involve long contexts and multiple transitions. Third, the benchmark is currently restricted to text and image modalities, excluding emerging audio and video inputs. 

Finally, while our teacher-forcing approach (using ground-truth history) successfully isolates the effects of task switching, it does not capture self-induced interference, which refers to the cascading effect whereby a model's own generation errors in prior turns corrupt subsequent context.
In real-world scenarios, a model's previous generation errors or hallucinations can propagate, likely causing more severe performance degradation than reported here.

\section*{Acknowledgements}
This paper is based on results obtained from AIST policy-based budget project "R\&D on Generative AI Foundation Models for the Physical Domain". We used ABCI 3.0 provided by AIST and AIST Solutions with support from “ABCI 3.0 Development Acceleration Use”.

\section*{Bibliographical References}\label{sec:reference}
\bibliographystyle{lrec2026-natbib}
\bibliography{lrec2026-example.bib}

@inproceedings{pang2005,
    title = {Seeing Stars: Exploiting Class Relationships for Sentiment Categorization with Respect to Rating Scales},
    author = {Bo Pang and Lillian Lee},
    editor = {Kevin Knight and Hwee Tou Ng and Kemal Oflazer},
    booktitle = {Proceedings of the 43rd Annual Meeting of the Association for Computational Linguistics (ACL 2005)},
    month = {June},
    year = {2005},
    address = {Ann Arbor, Michigan},
    publisher = {Association for Computational Linguistics},
    url = {https://aclanthology.org/P05-1015/},
    doi = {10.3115/1219840.1219855},
    pages = {115--124}
}

@inproceedings{hendrycks2021,
    title = {Measuring Massive Multitask Language Understanding},
    author = {Dan Hendrycks and Collin Burns and Steven Basart and Andy Zou and Mantas Mazeika and Dawn Song and Jacob Steinhardt},
    booktitle = {Proceedings of the 9th International Conference on Learning Representations (ICLR 2021)},
    year = {2021},
    publisher = {OpenReview.net},
    url = {https://huggingface.co/datasets/cais/mmlu}
}

@inproceedings{xiong2019tweetqa,
    title = {TWEETQA: A Social Media Focused Question Answering Dataset},
    author = {Wenhan Xiong and Jiawei Wu and Hong Wang and Vivek Kulkarni and Mo Yu and Shiyu Chang and Xiaoxiao Guo and William Yang Wang},
    editor = {Anna Korhonen and David Traum and Llu{\'i}s M{\`a}rquez},
    booktitle = {Proceedings of the 57th Annual Meeting of the Association for Computational Linguistics},
    month = {July},
    year = {2019},
    address = {Florence, Italy},
    publisher = {Association for Computational Linguistics},
    url = {https://aclanthology.org/P19-1496/},
    doi = {10.18653/v1/P19-1496},
    pages = {5020--5031}
}

@inproceedings{marino2019okvqa,
    author = {Kenneth Marino and Mohammad Rastegari and Ali Farhadi and Roozbeh Mottaghi},
    title = {OK-VQA: A Visual Question Answering Benchmark Requiring External Knowledge},
    booktitle = {Proceedings of the IEEE/CVF Conference on Computer Vision and Pattern Recognition (CVPR)},
    year = {2019},
    pages = {3195--3204},
    publisher = {Computer Vision Foundation / IEEE},
    doi = {10.1109/CVPR.2019.00331},
    url = {http://openaccess.thecvf.com/content_CVPR_2019/html/Marino_OK-VQA_A_Visual_Question_Answering_Benchmark_Requiring_External_Knowledge_CVPR_2019_paper.html}
}

@inproceedings{goyal2017vqa,
    author = {Yash Goyal and Tejas Khot and Douglas Summers-Stay and Dhruv Batra and Devi Parikh},
    title = {Making the V in VQA Matter: Elevating the Role of Image Understanding in Visual Question Answering},
    booktitle = {Proceedings of the IEEE Conference on Computer Vision and Pattern Recognition (CVPR)},
    year = {2017},
    pages = {6325--6334},
    publisher = {IEEE Computer Society},
    doi = {10.1109/CVPR.2017.670},
    url = {https://doi.org/10.1109/CVPR.2017.670}
}

@inproceedings{lin2014coco,
    author = {Tsung-Yi Lin and Michael Maire and Serge J. Belongie and James Hays and Pietro Perona and Deva Ramanan and Piotr Doll{\'a}r and C. Lawrence Zitnick},
    editor = {David J. Fleet and Tom{\'a}s Pajdla and Bernt Schiele and Tinne Tuytelaars},
    title = {Microsoft {COCO}: Common Objects in Context},
    booktitle = {Computer Vision -- ECCV 2014 -- 13th European Conference, Zurich, Switzerland, September 6-12, 2014, Proceedings, Part V},
    series = {Lecture Notes in Computer Science},
    volume = {8693},
    pages = {740--755},
    publisher = {Springer},
    year = {2014},
    url = {https://doi.org/10.1007/978-3-319-10602-1_48},
    doi = {10.1007/978-3-319-10602-1_48}
}

@inproceedings{gupta2024llm,
  title = {LLM Task Interference: An Initial Study on the Impact of Task-Switch in Conversational History},
  author = {Gupta, Akash and Sheth, Ivaxi and Raina, Vyas and Gales, Mark and Fritz, Mario},
  booktitle = {Proceedings of the 2024 Conference on Empirical Methods in Natural Language Processing (EMNLP)},
  pages = {14633--14652},
  year = {2024},
  url = {https://aclanthology.org/2024.emnlp-main.811/}
}

@article{brown2020language,
  title={Language models are few-shot learners},
  author={Brown, Tom and Mann, Benjamin and Ryder, Nick and Subbiah, Melanie and Kaplan, Jared and Dhariwal, Prafulla and Neelakantan, Arvind and Shyam, Pranav and Sastry, Girish and Askell, Amanda and others},
  journal={Advances in Neural Information Processing Systems},
  volume={33},
  pages={1877--1901},
  year={2020}
}

@article{openai2023gpt4,
  title={GPT-4 Technical Report},
  author={OpenAI},
  journal={arXiv preprint arXiv:2303.08774},
  year={2023}
}

@article{chen2023octavius,
  title = {Octavius: Mitigating Task Interference in MLLMs via LoRA-MoE},
  author = {Chen, Zeren and Wang, Ziqin and Wang, Zhen and Liu, Huayang and Yin, Zhenfei and Liu, Si and Sheng, Lu and Ouyang, Wanli and Qiao, Yu and Shao, Jing},
  journal = {arXiv preprint arXiv:2311.02684},
  year = {2023},
  url = {https://arxiv.org/abs/2311.02684}
}

@inproceedings{alayrac2022flamingo,
author = {Alayrac, Jean-Baptiste and Donahue, Jeff and Luc, Pauline and Miech, Antoine and Barr, Iain and Hasson, Yana and Lenc, Karel and Mensch, Arthur and Millicah, Katie and Reynolds, Malcolm and Ring, Roman and Rutherford, Eliza and Cabi, Serkan and Han, Tengda and Gong, Zhitao and Samangooei, Sina and Monteiro, Marianne and Menick, Jacob and Borgeaud, Sebastian and Brock, Andrew and Nematzadeh, Aida and Sharifzadeh, Sahand and Binkowski, Mikolaj and Barreira, Ricardo and Vinyals, Oriol and Zisserman, Andrew and Simonyan, Karen},
title = {Flamingo: a visual language model for few-shot learning},
year = {2022},
isbn = {9781713871088},
publisher = {Curran Associates Inc.},
address = {Red Hook, NY, USA},
booktitle = {Proceedings of the 36th International Conference on Neural Information Processing Systems},
articleno = {1723},
numpages = {21},
location = {New Orleans, LA, USA},
series = {NIPS '22}
}

@misc{agrawal2024pixtral12b,
      title={Pixtral 12B}, 
      author={Pravesh Agrawal and Szymon Antoniak and Emma Bou Hanna and Baptiste Bout and Devendra Chaplot and Jessica Chudnovsky and Diogo Costa and Baudouin De Monicault and Saurabh Garg and Theophile Gervet and Soham Ghosh and Amélie Héliou and Paul Jacob and Albert Q. Jiang and Kartik Khandelwal and Timothée Lacroix and Guillaume Lample and Diego Las Casas and Thibaut Lavril and Teven Le Scao and Andy Lo and William Marshall and Louis Martin and Arthur Mensch and Pavankumar Muddireddy and Valera Nemychnikova and Marie Pellat and Patrick Von Platen and Nikhil Raghuraman and Baptiste Rozière and Alexandre Sablayrolles and Lucile Saulnier and Romain Sauvestre and Wendy Shang and Roman Soletskyi and Lawrence Stewart and Pierre Stock and Joachim Studnia and Sandeep Subramanian and Sagar Vaze and Thomas Wang and Sophia Yang},
      year={2024},
      eprint={2410.07073},
      archivePrefix={arXiv},
      primaryClass={cs.CV},
      url={https://arxiv.org/abs/2410.07073}, 
}

@misc{yang2025qwen3technicalreport,
      title={Qwen3 Technical Report}, 
      author={An Yang and Anfeng Li and Baosong Yang and Beichen Zhang and Binyuan Hui and Bo Zheng and Bowen Yu and Chang Gao and Chengen Huang and Chenxu Lv and Chujie Zheng and Dayiheng Liu and Fan Zhou and Fei Huang and Feng Hu and Hao Ge and Haoran Wei and Huan Lin and Jialong Tang and Jian Yang and Jianhong Tu and Jianwei Zhang and Jianxin Yang and Jiaxi Yang and Jing Zhou and Jingren Zhou and Junyang Lin and Kai Dang and Keqin Bao and Kexin Yang and Le Yu and Lianghao Deng and Mei Li and Mingfeng Xue and Mingze Li and Pei Zhang and Peng Wang and Qin Zhu and Rui Men and Ruize Gao and Shixuan Liu and Shuang Luo and Tianhao Li and Tianyi Tang and Wenbiao Yin and Xingzhang Ren and Xinyu Wang and Xinyu Zhang and Xuancheng Ren and Yang Fan and Yang Su and Yichang Zhang and Yinger Zhang and Yu Wan and Yuqiong Liu and Zekun Wang and Zeyu Cui and Zhenru Zhang and Zhipeng Zhou and Zihan Qiu},
      year={2025},
      eprint={2505.09388},
      archivePrefix={arXiv},
      primaryClass={cs.CL},
      url={https://arxiv.org/abs/2505.09388}, 
}

@InProceedings{Vedantam_2015_CVPR,
author = {Vedantam, Ramakrishna and Lawrence Zitnick, C. and Parikh, Devi},
title = {CIDEr: Consensus-Based Image Description Evaluation},
booktitle = {Proceedings of the IEEE Conference on Computer Vision and Pattern Recognition (CVPR)},
month = {June},
year = {2015}
}

@article{shen2024multimodal,
  title={Multimodal Instruction Tuning with Conditional Mixture of LoRA},
  author={Shen, Ying and Xu, Zhiyang and Wang, Qifan and Cheng, Yu and Yin, Wenpeng and Huang, Lifu},
  journal={arXiv preprint arXiv:2402.15896},
  year={2024}
}

@misc{gemmateam2025gemma3technicalreport,
      title={Gemma 3 Technical Report}, 
      author={{Gemma Team}},
      year={2025},
      eprint={2503.19786},
      archivePrefix={arXiv},
      primaryClass={cs.CL},
      url={https://arxiv.org/abs/2503.19786}, 
}

@inproceedings{visual_instruction_tuning,
author = {Liu, Haotian and Li, Chunyuan and Wu, Qingyang and Lee, Yong Jae},
title = {Visual instruction tuning},
year = {2023},
publisher = {Curran Associates Inc.},
address = {Red Hook, NY, USA},
booktitle = {Proceedings of the 37th International Conference on Neural Information Processing Systems},
articleno = {1516},
numpages = {25},
location = {New Orleans, LA, USA},
series = {NIPS '23}
}

@inproceedings{kwon2023efficient,
  title={Efficient Memory Management for Large Language Model Serving with PagedAttention},
  author={Woosuk Kwon and Zhuohan Li and Siyuan Zhuang and Ying Sheng and Lianmin Zheng and Cody Hao Yu and Joseph E. Gonzalez and Hao Zhang and Ion Stoica},
  booktitle={Proceedings of the ACM SIGOPS 29th Symposium on Operating Systems Principles},
  year={2023}
}

@article{liu-etal-2024-lost,
    title = "Lost in the Middle: How Language Models Use Long Contexts",
    author = "Liu, Nelson F.  and
      Lin, Kevin  and
      Hewitt, John  and
      Paranjape, Ashwin  and
      Bevilacqua, Michele  and
      Petroni, Fabio  and
      Liang, Percy",
    journal = "Transactions of the Association for Computational Linguistics",
    volume = "12",
    year = "2024",
    address = "Cambridge, MA",
    publisher = "MIT Press",
    url = "https://aclanthology.org/2024.tacl-1.9/",
    doi = "10.1162/tacl_a_00638",
    pages = "157--173"
}

@misc{bai2025qwen3vltechnicalreport,
      title={Qwen3-VL Technical Report}, 
      author={Shuai Bai and Yuxuan Cai and Ruizhe Chen and Keqin Chen and Xionghui Chen and Zesen Cheng and Lianghao Deng and Wei Ding and Chang Gao and Chunjiang Ge and Wenbin Ge and Zhifang Guo and Qidong Huang and Jie Huang and Fei Huang and Binyuan Hui and Shutong Jiang and Zhaohai Li and Mingsheng Li and Mei Li and Kaixin Li and Zicheng Lin and Junyang Lin and Xuejing Liu and Jiawei Liu and Chenglong Liu and Yang Liu and Dayiheng Liu and Shixuan Liu and Dunjie Lu and Ruilin Luo and Chenxu Lv and Rui Men and Lingchen Meng and Xuancheng Ren and Xingzhang Ren and Sibo Song and Yuchong Sun and Jun Tang and Jianhong Tu and Jianqiang Wan and Peng Wang and Pengfei Wang and Qiuyue Wang and Yuxuan Wang and Tianbao Xie and Yiheng Xu and Haiyang Xu and Jin Xu and Zhibo Yang and Mingkun Yang and Jianxin Yang and An Yang and Bowen Yu and Fei Zhang and Hang Zhang and Xi Zhang and Bo Zheng and Humen Zhong and Jingren Zhou and Fan Zhou and Jing Zhou and Yuanzhi Zhu and Ke Zhu},
      year={2025},
      eprint={2511.21631},
      archivePrefix={arXiv},
      primaryClass={cs.CV},
      url={https://arxiv.org/abs/2511.21631}, 
}


\end{document}